\documentclass{article}
\usepackage[final,main]{neurips_2026}

\usepackage[utf8]{inputenc}
\usepackage[T1]{fontenc}
\usepackage{hyperref}
\hypersetup{
    colorlinks=true,
    linkcolor=blue,
    pdfpagemode=FullScreen,
        citecolor=blue,
    }
\usepackage{url}
\usepackage{booktabs}
\usepackage{amsfonts}
\usepackage{nicefrac}
\usepackage{microtype}
\usepackage{xcolor}

\usepackage{amsmath}
\usepackage{graphicx}
\usepackage{subcaption}
\usepackage{multirow}

\title{How Much is Left? LLMs Linearly Encode Their Remaining Output Length}

\author{%
  Mohamed Amine Merzouk\textsuperscript{\,1,2,$\ast$}\quad
  Dmitri Carpov\textsuperscript{\,3,$\ast$}\quad
  Mirko Bronzi\textsuperscript{\,3,$\ast$}\\[4pt]
  \bfseries Damiano Fornasiere\textsuperscript{\,3,$\diamond$}\quad
  Adam Oberman\textsuperscript{\,1,2,3,$\diamond$}\\[8pt]
  \textsuperscript{1}\,Mila -- Quebec AI Institute\quad
  \textsuperscript{2}\,McGill University\quad
  \textsuperscript{3}\,LawZero
}

\begin{document}

\maketitle
{\renewcommand{\thefootnote}{}\footnotetext{$^\ast$Joint first authors. \quad $^\diamond$Joint senior authors. \quad Correspondence to: \texttt{<mohamed-amine.merzouk@mila.quebec>}.}}

\begin{abstract}
Large Language models generate tokens, without explicitly keeping track of the length of the generated text.  This paper asks the question: can we estimate the number of tokens remaining in a generated response, using just a linear probe on the model's hidden states?  Previous work on inference scheduling established that length is statistically predictable, using nonlinear methods.   This paper contributes a linear representation method, using frozen hidden states of 7-8B instruction-tuned models, across seven datasets. We show that total completion length can be estimated, with performance stronger than statistical baselines, and stronger than the fair comparison to the prediction at the last token of the prompt. On the natural-language datasets, the probe's error lies between $0.57$ and $0.88$ of the statistical baseline's.  Interestingly, when a retraction token occurs (``Wait''), the probe predicted length jumps upward by over $70$ tokens, on average, compared to the baseline, where it drops by $0.7$, on average.  Moreover, the direction found can be used for length steering: adding to the probe direction increases completion length and vice-versa, on Llama-3.1-8B / GSM8K.  These results indicate that LLMs have a linearly decodable internal representation of output length. 
\begin{center}
    Code: \url{https://github.com/mamerzouk/llm-output-length}
 
\end{center}
\end{abstract}

\section{Introduction}\label{sec:intro}
In this paper, we are interested in predicting remaining  length of the output generated by an LLM, because it serves as a ground-truth proxy for indirectly addressing questions which are relevant to AI safety.  One such relevant question, is the commitment question: has a model committed to a particular course of action before generating a response? or is it changing its response as it generates. 
 The behaviour was recently  shown in the context of rhyming: models choose a rhyme word in advance of the line where it appears~\citep{lindsey2025biology}.  A retraction token that appears without an updward shift in the remaining length is a sign of decorative reasoning, ~\citep{tts2025}. Here we find that the model's hidden state allows us to accurately predict the remaining length.  (The remaining length of an answer generated by an LLM is the number of tokens left in the response before the end-of-sequence (EOS) token.)  Notably, the estimate of remaining length jumps upward at a retraction token, e.g., ``Wait'' (Section~\ref{sec:res:dynamic}), and the steering along the direction found allows us to control the length of a given answer (Section~\ref{sec:res:steering}).  

\begin{figure}[t]
\centering
\includegraphics[width=0.62\linewidth]{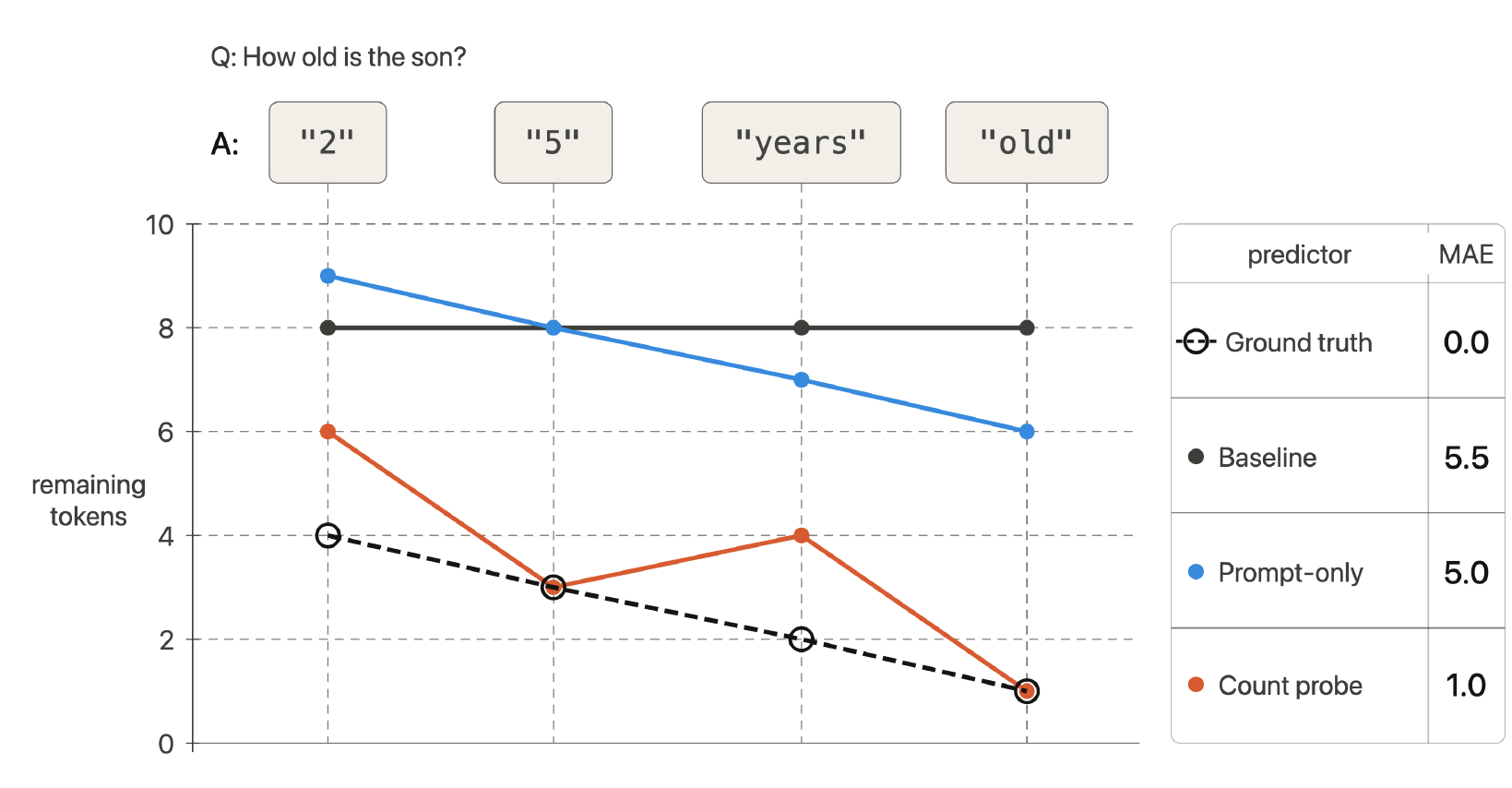}
\caption{The three predictors on a four-token answer to ``How old is the son?'', followed by EOS (ground-truth remaining counts 4, 3, 2, 1; dashed line, hollow markers). The legend's ``Baseline'', ``Prompt-only'' and ``Count probe'' are the constant-median baseline, the Completion Length Probe with exact countdown, and the Remaining Count Probe. The constant-median baseline outputs the dataset median at every position. The Completion Length Probe reads the prompt's last hidden state once and counts down by one per token, capturing the shape with a fixed offset (9, 8, 7, 6). The Remaining Count Probe reads the hidden state at every position; its predictions (6, 3, 4, 1) are not monotonic, yet it has the lowest MAE (mean absolute error, in tokens; lower is better): 1.0 against 5.0 and 5.5. Toy example for exposition.}
\label{fig:1}
\end{figure}

We train two linear probes~\citep{alain2017probes,belinkov2022probing} on the frozen model's residual stream.  This means that what the probe recovers is present in the state  (Section~\ref{subsec:probes}).  For a completion of length $T$, the Completion Length Probe probe is trained to predict, $T$, from the prompt's last state.  On the other hand, the Remaining Count Probe is trained at positions, $t$, to predict the remaining count, $r_t = T - t$, from the state, $h_t$.  In order to measure the value added by seeing multiple state vectors, we compare the Remaining Count Probe to a prediction defined by taking the  Completion Length Probe, and simply counting down by one at each token  (Figure~\ref{fig:1}).  We also compare the probe with the constant-median baseline, which corresponds to the statistically optimal value if a model was forced to output a single number on a dataset, using the same training loss (which is defined below).

Predicting prompt length has previously been of interest for batch scheduling, ~\citep{zheng2023response,qiu2024proxy,shahout2024trail}, and \citet{piotrowski2025tokens}, but there it used nonlinear predictors on a single corpus.  We design experiments to address the following questions: does a probe generalize from one dataset to another? Does its estimate respond when the model retracts? Does steering along its direction control the length of the response of the model?

\paragraph{Contributions.} We report that:
\begin{enumerate}
    \item \textbf{Transfer asymmetry.} Probes trained on natural-language data transfer to other datasets (beating the target dataset's own constant-median baseline), the synthetic ones included, far more often than synthetic-trained probes transfer to natural language (Section~\ref{app:cross-dataset}).
    \item \textbf{Prompt-end decodability.} Total completion length is linearly decodable from the prompt's last hidden state, before any token is generated, in every model and dataset cell reported (Section~\ref{sec:results}).
    \item \textbf{Per-token decodability.} In most cells, states in the middle of a completion add linearly decodable information beyond a countdown from the prompt-end estimate, as shown by comparing a probe read at every position with a constant baseline and with a countdown from that estimate (Section~\ref{sec:res:three-way}).
    \item \textbf{Retraction.} At a retraction token (``Wait''), the decoded estimate jumps upward relative to matched non-retraction positions (Qwen3-4B / GSM8K, Section~\ref{sec:res:dynamic}).
    \item \textbf{Steering.} Adding the probe direction to the residual stream at one layer lengthens or shortens completions monotonically (Llama-3.1-8B / GSM8K, layer 24, with the layer and range chosen after a three-layer sweep, Section~\ref{sec:res:steering}).
\end{enumerate}
These results are for naturally terminated completions - completions which did not terminate were excluded. 

\section{Setup}
\label{sec:methods}

\subsection{Target and indexing}
Let $M$ be a frozen autoregressive language model and $(x, y)$ a (prompt, completion) pair, with $y = (y_1, \ldots, y_T)$ generated by $M$ until it emits EOS at position $T$ or reaches a maximum-length cutoff. A pair \emph{terminates naturally} when EOS is emitted, and only naturally terminated pairs are used for training and evaluation. For each completion position $t \in \{1, \ldots, T\}$ the target is the remaining token count
\begin{equation}
r_t = T - t .
\end{equation}
For each position, prompt included, we extract the residual-stream state $h_t^{(\ell)} \in \mathbb{R}^d$ at every layer $\ell \in \{0, \ldots, L\}$ in one forward pass over the full sequence with $M$ frozen. $h_t^{(\ell)}$ is the state after reading token $t$. Causal masking excludes every token after $t$, so teacher-forced re-encoding of a generated sequence gives position-wise the same states as generation. $r_t$ counts the tokens emitted after position $t$ and never token $t$ itself. EOS is counted in $T$, so $r_t = 0$ exactly at the EOS position. A sequence that reaches the cutoff has no observed $T$, and assigning it a lower-bound target would bias the probe toward underestimation, so such sequences are masked entirely. The released extraction code enforces these conventions and tests them, and Appendix~\ref{app:exclusion} reports how many generations the filter removes.

\subsection{Probe family}
\label{subsec:probes}
A probe $f_\theta : \mathbb{R}^d \to \mathbb{R}$ is a single linear layer with no nonlinearity, so it cannot compute its target from weak features, which is the selectivity concern of \citet{hewitt2019control}. Unless stated otherwise, $h_t$ is the concatenation of the states at all layers. We compare three predictors of $r_t$ (Figure~\ref{fig:1}).

The \textbf{Remaining Count Probe} regresses $\hat{r}_t = f_\theta(h_t)$ at every completion position with squared loss. Its prediction depends on $h_t$ and not on $t$, so it need not decrease monotonically, which Section~\ref{sec:res:dynamic} uses.

The \textbf{Completion Length Probe} is trained only at the prompt's last position, with target $T$ and squared loss. Its prediction $\hat{T}_0$ is turned into a per-position predictor by the \textbf{exact countdown} $\hat{r}_t = \max(\hat{T}_0 - t - 1, 0)$, which uses no hidden state after the prompt. Under the stochastic decoding that generated the data, $T$ varies across resamples of the same prompt, so this probe estimates a central tendency of the length distribution and not the realized sample.

The \textbf{constant-median baseline} predicts the median of $r_t$ at every position. Each probe is trained with its own AdamW optimizer over three seeds.

\subsection{Baselines and metric}
\label{subsec:stat-baselines}
\label{subsec:evaluation}
Every regression result is reported as MAE in tokens, lower is better, averaged over the eval split with every completion position weighted equally. The Completion Length Probe is scored by its absolute error $|\hat{T}_0 - T|$ at the prompt's last position. Under absolute loss the best constant predictor is the median, so the reference is the constant-median baseline, fit on the eval split. It is the lowest-MAE constant on that split, which makes it the strictest constant reference, and its MAE is the mean absolute deviation of the eval distribution. A probe below it carries information about $r_t$ that no constant predictor has.

\subsection{Models and datasets}
\label{subsec:models-datasets}
\label{subsec:datasets}
The primary grid is three instruction-tuned 7--8B models, Llama-3.1-8B-Instruct~\citep{meta2024llama3}, Olmo-3-7B-Instruct~\citep{olmo2026olmo3} and Mistral-7B-Instruct-v0.3~\citep{jiang2023mistral}, which differ in tokenizer, pretraining data and instruction tuning. Two Qwen3 models~\citep{yang2025qwen3} go through the same pipeline in specific roles: Qwen3-4B for the retraction analysis of Section~\ref{sec:res:dynamic} and Qwen3-32B for a scale check. There are seven datasets. Count and Countdown are synthetic: the prompt asks the model to count up or down to $n \in [0, 300]$, so $T$ is fixed by the prompt. The other five are GSM8K~\citep{cobbe2021gsm8k}, MATH~\citep{hendrycks2021math}, MMLU-Pro~\citep{wang2024mmlupro}, OpenThoughts-1k~\citep{guha2025openthoughts} and TriviaQA~\citep{joshi2017triviaqa}, spanning short retrieval to long reasoning traces. Mistral-7B / TriviaQA was not run for compute reasons. We call a (model, dataset) pair a cell, and in the transfer matrices of Section~\ref{app:cross-dataset} a (training set, evaluation set) pair. Appendix~\ref{app:details} gives prompts, split sizes and hyperparameters.

\section{Remaining length is linearly decodable}
\label{sec:results}

Probe MAEs are means over three training seeds. In Tables~\ref{tab:prompt-only-count-mae} and~\ref{tab:prompt-only-static-decay-mae}, per-seed variation was below one token of MAE, smaller than the spread across cells, so per-seed error bars are omitted.

\begin{figure}[t]
\centering
\includegraphics[width=\linewidth]{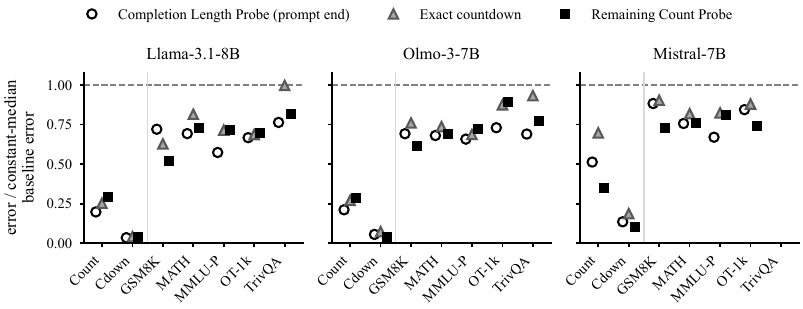}
\caption{Error of each predictor divided by the constant-median baseline's error in the same cell (lower is better; the dashed line at $1$ is the baseline), mean over three seeds, on naturally terminated completions. Completion Length Probe: prompt-end error $|\hat{T}_0 - T|$ (Table~\ref{tab:prompt-only-count-mae}). Exact countdown and Remaining Count Probe: per-token MAE (Table~\ref{tab:prompt-only-static-decay-mae}). The two synthetic datasets are left of the vertical line. Cdown: Countdown; MMLU-P: MMLU-Pro; OT-1k: OpenThoughts-1k; TrivQA: TriviaQA. Mistral-7B / TriviaQA was not run.}
\label{fig:relative-error}
\end{figure}

\paragraph{At the prompt's last position.}
\label{sec:res:headline}
Table~\ref{tab:prompt-only-count-mae} compares the Completion Length Probe with the constant-median baseline at the prompt's last position. The probe has the lower error in every cell reported. On Countdown, where $T$ is a function of the prompt, Llama's error is $5.27$ tokens against $150.18$. On the natural-language sets the probe's error is $0.57$ to $0.88$ of the baseline's (Figure~\ref{fig:relative-error}, Appendix~\ref{app:relative-mae}).

Because $T$ varies across resamples of the same prompt, a predictor that reads only the prompt cannot beat the spread of $T$ within a prompt. On Llama-3.1-8B / GSM8K we drew ten fresh completions per test prompt. Scored against the median of the other nine draws, the mean spread is $34.8$ tokens, and against the median of all ten, $30.0$. The probe's error is $1.22$ to $1.41$ times this spread, and it captures $69\%$ of the range between the constant-median baseline and the leave-one-out spread (Appendix~\ref{app:sampling-floor}).

On the same cell, a ridge regression on prompt length and TF-IDF features recovers about half of the probe's improvement over the constant-median baseline (Table~\ref{tab:controls}, Appendix~\ref{app:controls}).

\begin{table}[t]
\centering
\scriptsize
\caption{Prompt-end absolute error, in tokens (lower is better), of the Completion Length Probe against the constant-median baseline: $|\hat{T}_0 - T|$ at the prompt's last position, averaged over the eval split of naturally terminated completions, mean over three seeds. Bold marks the lower error per cell. Mistral-7B / TriviaQA was not run.}
\label{tab:prompt-only-count-mae}
\begin{tabular}{l rrrrrrr}
\toprule
Model & Count & Countdown & GSM8K & MATH & MMLU-Pro & OpenThoughts-1k & TriviaQA \\
\midrule
\multicolumn{8}{c}{Llama-3.1-8B} \\
\midrule
Constant median & 150.17 & 150.18 &  58.66 & 166.32 & 204.21 & 212.24 &  57.86 \\
Completion Length Probe & \textbf{29.73} & \textbf{5.27} & \textbf{42.29} & \textbf{115.29} & \textbf{117.19} & \textbf{141.65} & \textbf{44.20} \\
\midrule
\multicolumn{8}{c}{Olmo-3-7B} \\
\midrule
Constant median   & 147.67 & 150.58 & 116.17 & 194.05 & 204.82 & 272.58 & 160.56 \\
Completion Length Probe  & \textbf{31.22} & \textbf{8.40} & \textbf{80.46} & \textbf{132.13} & \textbf{134.86} & \textbf{199.13} & \textbf{110.84} \\
\midrule
\multicolumn{8}{c}{Mistral-7B} \\
\midrule
Constant median & 263.15 & 265.12 &  84.81 & 174.71 & 196.16 & 195.77 & --     \\
Completion Length Probe  & \textbf{135.09} & \textbf{35.92} & \textbf{74.92} & \textbf{132.28} & \textbf{131.51} & \textbf{165.36} & --     \\
\bottomrule
\end{tabular}
\end{table}

\paragraph{During generation.}
\label{sec:res:three-way}
Table~\ref{tab:prompt-only-static-decay-mae} compares the Remaining Count Probe with the constant-median baseline and the exact countdown at every completion position. The countdown receives $t$ as an explicit input, and its accuracy is inherited from the prompt-end estimate. Where the probe is below the constant-median baseline, the hidden state carries information about $r_t$ that no constant predictor has. Where it is also below the countdown, states in the middle of the completion add linearly readable information beyond what the countdown from the prompt-end estimate provides, so the estimate is revised during generation. Where the two are within a token of each other, the per-seed variation, the prompt-end estimate suffices for that cell. The probe is below the constant-median baseline in all $20$ cells (Figure~\ref{fig:relative-error}), below the countdown by more than a token in $13$, including every Mistral cell, and within a token of it in two (Llama / Countdown and Llama / MMLU-Pro). On Llama-3.1-8B / GSM8K, a probe of the same architecture and budget trained on shuffled targets (across the batch, or within each completion), following the control tasks of \citet{hewitt2019control}, does far worse than the probe trained in the same run (Table~\ref{tab:controls}, Appendix~\ref{app:controls}), so the accuracy depends on the true pairing of states and targets.

\paragraph{Scale.} On Qwen3-32B / GSM8K (one seed), the Remaining Count Probe reaches $32.8$ against the constant-median baseline's $85.8$ (ratio $0.38$; Llama-3.1-8B $0.52$) and the Completion Length Probe $49.5$ against $69.9$ ($0.71$; Llama $0.72$; Appendix~\ref{app:controls}).

\begin{table}[t]
\centering
\scriptsize
\caption{Per-token MAE, in tokens (lower is better), averaged over every completion position of the eval split, mean over three seeds. The constant median predicts the same value at every position; the exact countdown starts from the prompt-end estimate $\hat{T}_0$ and decrements by one per token; the Remaining Count Probe reads $h_t$ at every position. Bold marks the lowest error per cell. Mistral-7B / TriviaQA was not run.}
\label{tab:prompt-only-static-decay-mae}
\begin{tabular}{l rrrrrrr}
\toprule
Model & Count & Countdown & GSM8K & MATH & MMLU-Pro & OpenThoughts-1k & TriviaQA \\
\midrule
\multicolumn{8}{c}{Llama-3.1-8B} \\
\midrule
Constant median & 117.81 & 117.82 &  70.15 & 151.20 & 172.49 & 187.83 &  61.09 \\
Exact countdown & \textbf{29.80} &   4.82 & 44.04 & 123.32 & \textbf{123.31} & \textbf{129.09} &  60.94 \\
Remaining Count Probe         &  34.41 & \textbf{4.38} & \textbf{36.59} & \textbf{109.87} & 123.56 & 131.47 & \textbf{50.05} \\
\midrule
\multicolumn{8}{c}{Olmo-3-7B} \\
\midrule
Constant median & 116.54 & 117.84 & 124.32 & 178.83 & 184.82 & 219.01 & 151.00 \\
Exact countdown & \textbf{31.59} &   8.64 & 94.51 & 131.87 & \textbf{127.19} & \textbf{191.75} & 140.88 \\
Remaining Count Probe         &  33.36 & \textbf{4.50} & \textbf{76.65} & \textbf{123.10} & 133.54 & 195.97 & \textbf{116.75} \\
\midrule
\multicolumn{8}{c}{Mistral-7B} \\
\midrule
Constant median & 197.45 & 200.23 &  91.67 & 161.41 & 171.10 & 179.71 & --     \\
Exact countdown   & 137.77 &  37.36 & 82.90 & 132.19 & 141.05 & 158.16 & --     \\
Remaining Count Probe         & \textbf{69.51} & \textbf{20.81} & \textbf{66.94} & \textbf{122.23} & \textbf{138.55} & \textbf{133.63} & --     \\
\bottomrule
\end{tabular}
\end{table}

\begin{table}[t]
\centering
\small
\caption{Controls on Llama-3.1-8B / GSM8K, eval MAE in tokens (lower is better). Left: prompt-end prediction of $T$. Right: per-token prediction of $r_t$, with the Remaining Count Probe trained in the same run as the shuffled heads.}
\label{tab:controls}
\begin{tabular}{lr@{\qquad}lr}
\toprule
\multicolumn{2}{c}{Prompt end (predict $T$)} & \multicolumn{2}{c}{Per token (predict $r_t$)} \\
\midrule
constant median & 58.7 & constant median & 70.2 \\
ridge: length features only & 52.4 & shuffled labels (global) & 66.5 \\
ridge: TF-IDF only & 52.0 & shuffled labels (within completion) & 67.6 \\
ridge: all surface features & 50.5 & & \\
Completion Length Probe & \textbf{42.3} & Remaining Count Probe & \textbf{34.5} \\
\bottomrule
\end{tabular}
\end{table}

\section{Transfer across datasets is asymmetric}
\label{app:cross-dataset}

A probe that only fits its training set's distribution of $T$ should fail on a dataset whose distribution differs, so transfer tests whether the decoded direction carries length beyond the training distribution. For each model we train a Remaining Count Probe on each training dataset of its matrix and evaluate it on each evaluation dataset. Each off-diagonal entry is judged against the \emph{eval-median floor}, the MAE of the evaluation dataset's own constant-median baseline fit on its eval split. Table~\ref{tab:cross-llama-abs} shows Llama-3.1-8B. The Mistral-7B matrix (four training sets, five evaluation sets) and the Olmo-3-7B matrix (all seven) are in Appendix~\ref{app:transfer}.

Pooled over the three models, probes trained on a natural-language dataset beat the evaluation dataset's own floor on $29$ of $44$ off-diagonal cells, against $2$ of $28$ for synthetic-trained probes, both of which are Count to Countdown. Across domains the split is $10$ of $18$ natural-to-synthetic against $0$ of $22$ synthetic-to-natural. Per model, natural-trained probes win $7$ of $12$, $4$ of $8$ and $18$ of $24$ cells (Llama, Mistral, Olmo). One-sided Fisher exact tests give $p = 4.1 \times 10^{-7}$ pooled and $p = 5.2 \times 10^{-5}$ across domains. Cells sharing a training row are not independent, so we quote these as summaries beside the per-model counts.

On Llama, a Count-trained probe scores $228.93$ on MMLU-Pro against an eval-median floor of $172.49$, while the OpenThoughts-1k-trained probe beats the floor on both synthetic sets and on MMLU-Pro. GSM8K is the hardest target: natural-trained probes beat its floor on $2$ of $6$ cells, against $27$ of $38$ for the other targets, and on Llama three of the five failures are GSM8K-trained probes. Appendix~\ref{app:transfer} gives the per-cell ratios and a regularization reading of the asymmetry.

\begin{table}[t]
\scriptsize
\centering
\caption{Cross-dataset per-token MAE of the Remaining Count Probe on Llama-3.1-8B, in tokens (lower is better), mean $\pm$ standard deviation over three training seeds (the MMLU-Pro-trained row has one seed). Rows: training dataset; columns: evaluation dataset. The eval-median floor is the MAE of the evaluation dataset's own constant-median baseline. Bold marks off-diagonal entries below the floor. MATH and TriviaQA are not in the Llama matrix.}
\label{tab:cross-llama-abs}
\label{tab:llama-cross}
\begin{tabular}{l rr | rrr}
\toprule
Train & Count & Countdown & GSM8K & MMLU-Pro & OpenThoughts-1k \\
\midrule
\textit{eval-median floor} & 117.81 & 117.82 & 70.15 & 172.49 & 187.83 \\
\midrule
Count                  &  35.07 $\pm$ 0.43 &  \textbf{99.17} $\pm$ 1.09 & 101.21 $\pm$ 0.49 & 228.93 $\pm$ 0.53 & 280.02 $\pm$ 0.73 \\
Countdown              & 183.81 $\pm$ 0.14 &   4.13 $\pm$ 0.17 &  87.84 $\pm$ 0.19 & 236.40 $\pm$ 0.30 & 278.57 $\pm$ 0.32 \\
\midrule
GSM8K                  & 130.91 $\pm$ 3.57 & 149.68 $\pm$ 4.48 &  38.79 $\pm$ 0.44 & \textbf{164.27} $\pm$ 1.28 & 199.74 $\pm$ 1.33 \\
MMLU-Pro               & \textbf{105.01} & 118.96 &  \textbf{63.57} & 121.51 & \textbf{147.25} \\
OpenThoughts-1k        & \textbf{100.37} $\pm$ 2.01 & \textbf{101.03} $\pm$ 0.61 & 107.94 $\pm$ 3.59 & \textbf{133.05} $\pm$ 0.55 & 136.69 $\pm$ 0.41 \\
\bottomrule
\end{tabular}
\end{table}

\section{The estimate jumps at retraction tokens}
\label{sec:res:dynamic}

\begin{figure}[t]
\centering
\begin{subfigure}[t]{0.38\linewidth}
\centering
\includegraphics[width=\linewidth]{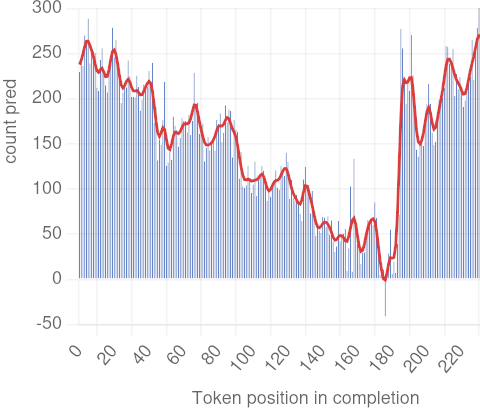}
\caption{Curated illustration: $\hat{r}_t$ on one high-error completion, around a retraction (positions 0 to 230).}
\end{subfigure}
\hfill
\begin{subfigure}[t]{0.6\linewidth}
\centering
\includegraphics[width=\linewidth,trim=0 0 0 34,clip]{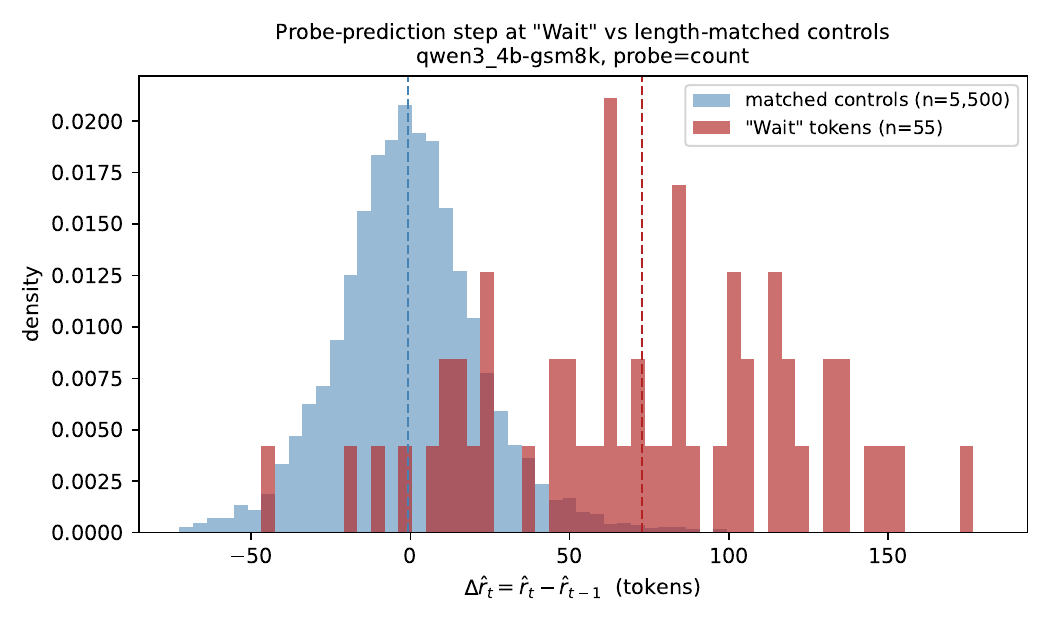}
\caption{Aggregate test: one-step deltas $\Delta\hat{r}_t$ at the $55$ ``Wait'' tokens against $5{,}500$ matched controls.}
\end{subfigure}
\caption{The Remaining Count Probe's estimate at retraction tokens (Qwen3-4B on GSM8K). (a) is a curated example from the high-error region of the eval set, chosen to illustrate the direction of the update; its absolute predictions are far from the truth. (b) is the evidence: deltas at every ``Wait'' token in the completions against non-retraction tokens matched on position bucket and per-completion MAE quartile. Retraction deltas center at $+72.8$, controls at $-0.7$.}
\label{fig:retraction}
\end{figure}

The constant-median baseline and the exact countdown never increase from one position to the next. The Remaining Count Probe reads $h_t$, so its prediction can rise. Figure~\ref{fig:retraction}a is a curated illustration: on a long, high-error completion, the prediction rises from $71$ to $277$ at the token ``Wait'' while the true remaining count keeps decreasing, and the absolute predictions stay far below the true count, which is about $800$ at the retraction, (Appendix~\ref{app:retraction}).

We use Qwen3-4B because retraction follows errors and the 7--8B models rarely err on GSM8K: ``Wait'' never occurs in Llama-3.1-8B's $1{,}313$ GSM8K evaluation completions and occurs three times in Qwen3-32B's. To measure this in aggregate, we test the token ``Wait'', matched exactly as a decoded token, against matched non-retraction positions on all $1{,}287$ Qwen3-4B / GSM8K evaluation completions, with no selection by error. This gives $55$ events in $45$ completions. A scan of all token types over the same completions ranks ``Wait'' first by mean one-step change (Appendix~\ref{app:retraction}). At each event we take $\Delta\hat{r}_t = \hat{r}_t - \hat{r}_{t-1}$ and compare it with the same quantity at $100$ non-retraction tokens per event from other completions, matched on position within the completion in 100-token buckets and on per-completion MAE quartile, excluding positions within 10 tokens of any occurrence of the trigger. Retraction deltas average $+72.8$ (median $+73.0$) against $-0.7$ (median $-1.1$) for controls, and $95\%$ of retraction steps move upward against $48\%$ of control steps (Cohen's $d = 3.10$). A sign-flip permutation test over per-completion paired differences, which respects clustering within completions, gives $p < 10^{-4}$ (Figure~\ref{fig:retraction}b). All $55$ events fall in the two worst MAE quartiles, and within the worst quartile, against controls from that quartile, $d = 3.12$.

The result concerns the token: the matching does not condition on token identity, so it does not separate an estimate that rises because more work is coming from a probe weight on ``Wait'' itself (Section~\ref{sec:limitations}).

\section{Steering along the direction changes length}
\label{sec:res:steering}

A decodable direction need not affect generation, so we test whether moving the residual stream along it changes length. The Remaining Count Probe reads all layers concatenated, so its weight vector splits into one slice per layer. For layer $\ell$, $w$ is that layer's slice of the probe trained on Llama-3.1-8B / GSM8K, and during generation we add $\alpha\, w / \lVert w \rVert$ to the residual stream at layer $\ell$ at every position. We run Llama-3.1-8B on the full GSM8K test set ($n = 1{,}319$) at layers 8, 16 and 24 for $\alpha$ from $-12$ to $+12$, with the same prompts and sampling seed in every arm, so comparisons are paired per prompt.

\begin{figure}[t]
\centering
\includegraphics[width=0.9\linewidth]{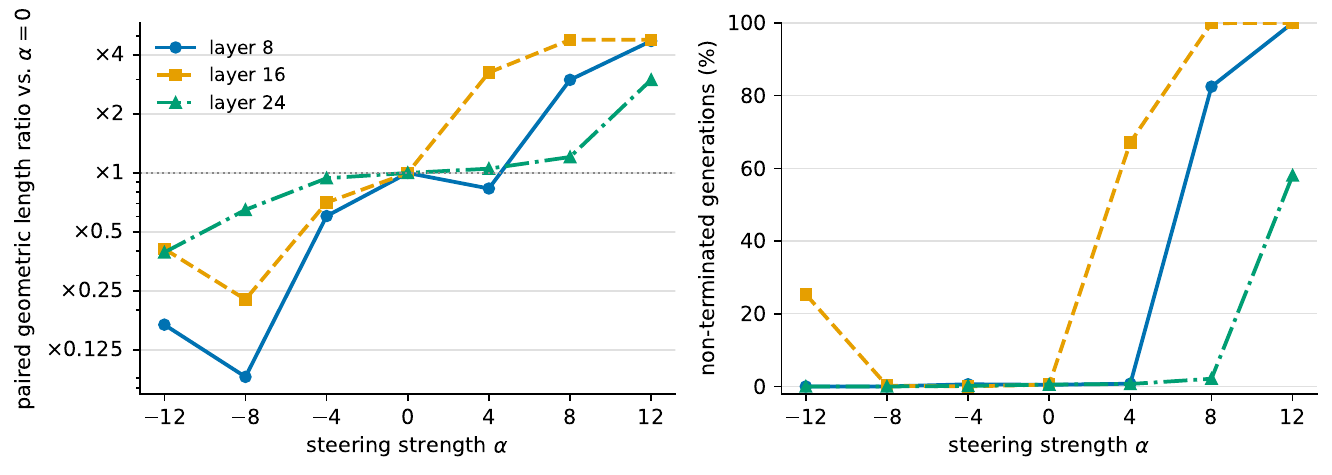}
\caption{Steering along the probe direction at layers 8, 16 and 24 (Llama-3.1-8B / GSM8K, $n = 1{,}319$ paired prompts per arm). Left: geometric mean of per-prompt length ratios against the $\alpha = 0$ arm (log scale). Right: fraction of generations reaching the 1,024-token cutoff. Layer 24 is monotone with few cutoffs over $\alpha \in [-8, +8]$. Layers 8 and 16 are not monotone and reach the cutoff at smaller $|\alpha|$.}
\label{fig:steering}
\end{figure}

At layer 24, over $\alpha$ from $-8$ to $+8$, the geometric mean of per-prompt length ratios changes monotonically from $0.65$ to $1.21$, and mean length from $155$ to $283$ tokens against $227$ unsteered (Figure~\ref{fig:steering}). Every arm differs from $\alpha = 0$ under a paired sign-flip test at $p < 10^{-4}$. No correction for the $18$ arms was applied, and every arm remains significant under a Bonferroni correction over them. We measured quality as the negative log-likelihood (NLL) of the steered completion under the unsteered model and as the rate of generations reaching the cutoff: over this range the NLL is at most $0.31$ against $0.18$ unsteered, and at most $2.1\%$ of generations reach the cutoff. Task accuracy under steering was not measured. Layer 24 and the range $\pm 8$ were chosen after inspecting the sweep.

At layers 8 and 16, $\alpha = -4$ is the only arm that moves length in the intended direction with NLL below $0.6$ and a cutoff rate at the natural level (NLL $0.58$ and $0.30$; cutoffs $0.6\%$ and $0.0\%$); the arms at $-8$ and $-12$ have length ratios below one but NLL $1.7$ to $5.1$, and at layer 16, $\alpha = -12$ sends $25\%$ of generations to the cutoff. At $\alpha = +4$, completions get shorter at layer 8, and most generations reach the cutoff at layer 16 (Appendix~\ref{app:steering-tables}).

\section{Related work}
\label{sec:related}

\paragraph{Predicting output length.}
\citet{zheng2023response} have the model state its expected response length and schedule batches on it; \citet{qiu2024proxy} predict length from the prompt with a small proxy model; \citet{shahout2024trail} classify remaining length into bins from intermediate-layer embeddings with a nonlinear head. Closest to us, \citet{piotrowski2025tokens} regress the per-position remaining count $r_t$ from frozen hidden states of three 7--8B instruction-tuned models, with an MLP over aggregated layers and a graph network over layers, on 1{,}000 Alpaca prompts. That work establishes predictability. Our probe is a single linear layer, so its accuracy shows that the signal is linearly available rather than computed by the probe. We report explicit constant-median and countdown baselines, without which raw error cannot show signal beyond the dataset's length distribution. We test transfer over seven datasets and three models, where prior work trains and tests on one corpus, and we add an aggregate retraction test and a steering test of the decoded direction.

\paragraph{Linear probes.}
Linear probes on frozen activations go back to \citet{alain2017probes} and recover linguistic structure~\citep{tenney2019bertpipeline,hewitt2019structuralprobe}, beliefs about truth~\citep{burns2023ccs,marks2024geometry}, world state~\citep{li2023othellogpt,nanda2023emergentlinear} and safety-relevant behavior~\citep{arditi2024refusal,zou2023repe}, with the linear representation hypothesis~\citep{elhage2022superposition,park2024lrh} accounting for when this is expected. Probe accuracy shows what is decodable, not what the model uses~\citep{belinkov2022probing}. We know of no out-of-distribution evaluation of a continuous probe target on the residual stream, and our cross-dataset matrices contribute one.

\paragraph{Commitment ahead of time and retractions.}
A single hidden state predicts several later tokens~\citep{pal2023future}, and circuit tracing shows a model choosing a rhyme word before writing the line~\citep{lindsey2025biology}. The literature calls this planning. We avoid the word ``plan'' for our results, which do not show that the model consults the estimate (Section~\ref{sec:limitations}). Shifts around the ``aha moment'' of DeepSeek-R1~\citep{guo2025deepseekr1} have been read skeptically~\citep{daliberti2026illusion,liu2025r1zerocritical,huang2024selfcorrect} and mechanistically~\citep{yang2025aha,tts2025,reasoningtheater2026}. Closest in method, \citet{reasoningtheater2026} probe answer confidence at every reasoning step. Inserting ``Wait'' is known to extend test-time computation~\citep{muennighoff2025s1}.

\paragraph{Exact counting.}
Transformers provably struggle with exact in-context counting~\citep{yehudai2025counting,barbero2024oversquashing,singh2024tokenizationcounts}. Our target is an approximate regression on the model's own future output, to which those results do not apply.

\section{Limitations}
\label{sec:limitations}

Steering shows the direction can be used to control length, not that the unperturbed model consults it. That would take activation patching or ablation of natural activity along the direction. There is also no random-direction steering control, so specificity to the probe direction is untested. The retraction result is measured on one cell, Qwen3-4B / GSM8K, for one token, ``Wait''. Its matching does not condition on token identity, so a probe weight on ``Wait'' could produce the same jump. The primary grid is 7--8B instruction-tuned models. Base models and frontier-scale models are future work. All results condition on natural termination. The filter removes $0$ to $0.5\%$ of generations on the synthetic sets and GSM8K, $4.8$ to $14\%$ on MMLU-Pro, $14$ to $16\%$ on MATH and $31$ to $37.5\%$ on OpenThoughts-1k (Llama-3.1-8B; Appendix~\ref{app:exclusion}). Mistral-7B / TriviaQA was not run, and the Llama and Mistral transfer matrices cover five datasets; Appendix~\ref{app:limitations-full} expands on each of these limitations.

\section{Conclusion}
\label{sec:conclusion}

The hidden state of an instruction-tuned LLM carries a linearly decodable estimate of how much of its own response remains. The estimate is present before generation begins, is revised during it in most cells, transfers from natural-language to synthetic data but not the reverse, and jumps at the retraction token tested (Section~\ref{sec:res:dynamic}). In the one cell tested, adding its direction to the residual stream changes how long the model writes (Section~\ref{sec:res:steering}). Whether the model's own computation reads the estimate is the open question. The retraction contrast is the closest existing test of it, and the starting point for the early-termination and decorative-reasoning uses of Section~\ref{sec:intro}.

\begin{ack}
D.F. conceived the project and supervised its initial phase. D.C. and M.B. developed the original codebase on which this work is built and ran the initial prototype experiments. M.A.M. extended the codebase, designed and ran the experiments reported in the paper, and contributed to writing. A.O. supervised the second stage of the project and wrote the manuscript. All authors reviewed and approved the final version.

This work was supported by the \href{https://www.nserc-crsng.gc.ca}{Natural Sciences and Engineering Research Council of Canada} (NSERC), by \href{https://www.coefficientgiving.org}{Coefficient Giving}, and by a \href{https://cifar.ca/ai/canada-cifar-ai-chairs/}{Canada CIFAR AI Chair}.
This research was enabled in part by support provided by \href{https://www.calculquebec.ca}{Calcul Qu\'ebec} and the \href{https://alliancecan.ca}{Digital Research Alliance of Canada}.

The authors used a large language model (Claude, Anthropic) to help draft, edit and format the text, and they reviewed all of it and take responsibility for its content.
\end{ack}

\newpage

\bibliography{bibliography}
\bibliographystyle{plainnat}
\newpage

\appendix

\section{Experimental details}
\label{app:details}
\label{app:hyperparams}

\paragraph{Datasets.} Count and Countdown use the prompts \texttt{"Count from 0 to \{n\}. Only output the numbers separated by a space. Start now:"} and \texttt{"Count down from \{n\} to 0. Only output the numbers separated by a space. Start now:"}, with $n \in [0, 300]$, so the completion length is fixed by the prompt and a probe that performs well on them reads the information from the hidden state. GSM8K is grade-school math, MATH competition math, MMLU-Pro multiple choice with reasoning, OpenThoughts-1k long reasoning traces and TriviaQA short retrieval. Each input is a chat-template message with a fixed system prompt (Table~\ref{tab:prompt-templates}) and a user message taken unmodified from the dataset. Table~\ref{tab:dataset-sizes} gives the split sizes.

\paragraph{Training.} Hidden states for both splits are extracted once per example and cached, and every probe reads the cache. All predictors share one forward pass and one dataloader per minibatch. For each minibatch, each active probe computes its loss, runs its backward pass and steps its own AdamW optimizer; the base model is never updated. Table~\ref{tab:hyperparams} lists the settings. The released repository (\url{https://github.com/mamerzouk/llm_output_length}) reproduces every configuration, and each run records its \texttt{config.yaml}.

\begin{table}[h]
\centering
\scriptsize
\caption{Probe-training hyperparameters, from the released code. In the code, \texttt{count} is the Remaining Count Probe and \texttt{prompt\_only\_count} the Completion Length Probe.}
\label{tab:hyperparams}
\begin{tabular}{l l l}
\toprule
Setting & Value & Notes \\
\midrule
Optimizer (per probe head) & AdamW & one optimizer per active probe head \\
Learning rate & $2 \times 10^{-4}$ & shared across probe heads \\
Weight decay & $0.01$ & \\
Max gradient norm & $10.0$ & gradient clipping \\
Loss (regression) & MSE & for \texttt{count}, \texttt{prompt\_only\_count} \\
LR scheduler & \texttt{ReduceLROnPlateau} & monitor \texttt{eval\_mae}, factor $0.5$, patience $5$ eval steps, $\min$\,LR $= 10^{-7}$ \\
Max steps & $4000$ & \\
Batch size (per device) & $8$ & training and eval; same across probes \\
Gradient accumulation & $1$ & \\
Eval cadence & every $100$ steps & \texttt{ReduceLROnPlateau} consumes \texttt{eval\_mae} from these evals \\
Probe input layer & all layers concatenated & \texttt{data.hidden\_layer = "all"} \\
Independent seeds & $\{0, 1, 2\}$ & seeds threaded through Torch / NumPy / Python / HF Trainer / dataloader \\
Hidden-state extraction & frozen forward pass & \texttt{torch.no\_grad()} over the full (prompt, completion) sequence \\
Generation (extraction) & \texttt{max\_new\_tokens} $=1024$ & \texttt{do\_sample=True}, $T = 0.7$, top-$p = 0.8$, top-$k = 20$ \\
Naturally terminated filter & \texttt{has\_eos = True} only & sequences that hit \texttt{max\_new\_tokens} excluded \\
\bottomrule
\end{tabular}
\end{table}

\begin{table}[h]
\centering
\footnotesize
\caption{Configured train and eval split sizes per dataset. The samples used by each probe are those that terminated naturally, so effective counts vary slightly by model.}
\label{tab:dataset-sizes}
\begin{tabular}{l rr l}
\toprule
Dataset & Train & Eval & Source and split \\
\midrule
Count       & 301        & 301        & synthetic, $n \in \{0, \ldots, 300\}$, one completion per length \\
Countdown   & 301        & 301        & synthetic, $n \in \{0, \ldots, 300\}$, one completion per length \\
GSM8K       & $7{,}473$  & $1{,}319$  & full \texttt{train} / \texttt{test} splits \\
MATH        & $7{,}500$  & $5{,}000$  & full \texttt{train} / \texttt{test} splits, concatenated over the seven subject configs \\
MMLU-Pro    & $10{,}000$ & $2{,}256$  & first $10{,}000$ for train, last $2{,}256$ for eval \\
OpenThoughts-1k & $800$  & $200$      & first $800$ / last $200$ of the $1$k sample \\
TriviaQA    & $10{,}000$ & $2{,}000$  & first $10{,}000$ for train, last $2{,}000$ for eval \\
\bottomrule
\end{tabular}
\end{table}

\begin{table}[h]
\centering
\small
\caption{System prompt and user-message source for each dataset. For the natural-language datasets the user message is the named HuggingFace field, unmodified.}
\label{tab:prompt-templates}
\label{app:prompts}
\begin{tabular}{l p{7.2cm} l}
\toprule
Dataset & System prompt & User message source \\
\midrule
Count           & ``You are a helpful assistant that follows instructions precisely. When asked to generate tokens, you produce exactly the requested number of tokens.'' & template (above) \\
Countdown       & ``You are a helpful assistant that follows instructions precisely. When asked to generate tokens, you produce exactly the requested number of tokens.'' & template (above) \\
GSM8K           & ``You are a helpful math tutor that solves grade school math problems step by step.'' & \texttt{question} field \\
MATH            & ``You are a helpful math tutor that solves math problems step by step.'' & \texttt{problem} field \\
MMLU-Pro        & ``You are a helpful assistant that solves problems step by step.'' & \texttt{question} field \\
OpenThoughts-1k & ``You are a helpful assistant that solves problems step by step.'' & \texttt{problem} field \\
TriviaQA        & ``You are a helpful assistant that solves problems step by step.'' & \texttt{question} field \\
\bottomrule
\end{tabular}
\end{table}

\paragraph{Exclusion rates.}
\label{app:exclusion}
Table~\ref{tab:exclusion} gives the fraction of generations removed by the naturally terminated filter on Llama-3.1-8B. Every exclusion is a cutoff at $1{,}024$ tokens, a budget set for compute reasons; no completion is empty. The TriviaQA rate was not recorded, and the Olmo and Mistral extraction metadata was not retained, so their rates would require regenerating the caches.

\begin{table}[h]
\centering
\small
\caption{Fraction of generations excluded by the naturally terminated filter (Llama-3.1-8B), by dataset and split. TriviaQA was not recorded.}
\label{tab:exclusion}
\begin{tabular}{l rrrrrr}
\toprule
Split & Count & Countdown & GSM8K & MMLU-Pro & MATH & OpenThoughts-1k \\
\midrule
train & 0\% & 0\% & 0.29\% & 4.79\% & 14.56\% & 37.50\% \\
eval  & 0\% & 0\% & 0.45\% & 14.01\% & 16.03\% & 31.00\% \\
\bottomrule
\end{tabular}
\end{table}

\section{Further decodability results}

\subsection{Error relative to the constant-median baseline}
\label{app:relative-mae}

Tables~\ref{tab:prompt-only-count-mae-rel} and~\ref{tab:prompt-only-static-decay-mae-rel} divide each entry of Tables~\ref{tab:prompt-only-count-mae} and~\ref{tab:prompt-only-static-decay-mae} by the constant-median baseline's error for the same cell, so values below $1$ beat the baseline and cells with different length scales can be compared. The two probes' ratios lie between $0.52$ and $0.89$ on the natural-language datasets and between $0.04$ and $0.51$ on Count and Countdown. On Llama-3.1-8B / TriviaQA the exact countdown ties the constant-median baseline ($1.00$), and the Remaining Count Probe reaches $0.82$.

\begin{table}[h!]
\centering
\scriptsize
\caption{Prompt-end error of the Completion Length Probe divided by the constant-median baseline's (Table~\ref{tab:prompt-only-count-mae}). Lower is better; below $1$ beats the baseline. Mistral-7B / TriviaQA was not run.}
\label{tab:prompt-only-count-mae-rel}
\begin{tabular}{l rrrrrrr}
\toprule
Model & Count & Countdown & GSM8K & MATH & MMLU-Pro & OpenThoughts-1k & TriviaQA \\
\midrule
Llama-3.1-8B & 0.20 & 0.04 & 0.72 & 0.69 & 0.57 & 0.67 & 0.76 \\
Olmo-3-7B    & 0.21 & 0.06 & 0.69 & 0.68 & 0.66 & 0.73 & 0.69 \\
Mistral-7B   & 0.51 & 0.14 & 0.88 & 0.76 & 0.67 & 0.84 & --   \\
\bottomrule
\end{tabular}
\end{table}

\begin{table}[h!]
\centering
\scriptsize
\caption{Per-token MAE divided by the constant-median baseline's (Table~\ref{tab:prompt-only-static-decay-mae}). Lower is better; below $1$ beats the baseline. Mistral-7B / TriviaQA was not run.}
\label{tab:prompt-only-static-decay-mae-rel}
\begin{tabular}{l rrrrrrr}
\toprule
Model & Count & Countdown & GSM8K & MATH & MMLU-Pro & OpenThoughts-1k & TriviaQA \\
\midrule
\multicolumn{8}{c}{Llama-3.1-8B} \\
\midrule
Exact countdown       & 0.25 & 0.04 & 0.63 & 0.82 & 0.71 & 0.69 & 1.00 \\
Remaining Count Probe & 0.29 & 0.04 & 0.52 & 0.73 & 0.72 & 0.70 & 0.82 \\
\midrule
\multicolumn{8}{c}{Olmo-3-7B} \\
\midrule
Exact countdown       & 0.27 & 0.07 & 0.76 & 0.74 & 0.69 & 0.88 & 0.93 \\
Remaining Count Probe & 0.29 & 0.04 & 0.62 & 0.69 & 0.72 & 0.89 & 0.77 \\
\midrule
\multicolumn{8}{c}{Mistral-7B} \\
\midrule
Exact countdown       & 0.70 & 0.19 & 0.90 & 0.82 & 0.82 & 0.88 & --   \\
Remaining Count Probe & 0.35 & 0.10 & 0.73 & 0.76 & 0.81 & 0.74 & --   \\
\bottomrule
\end{tabular}
\end{table}

\subsection{Per-layer probe sweep}
\label{app:per-layer}

\begin{figure}[h]
\centering
\includegraphics[width=0.75\linewidth]{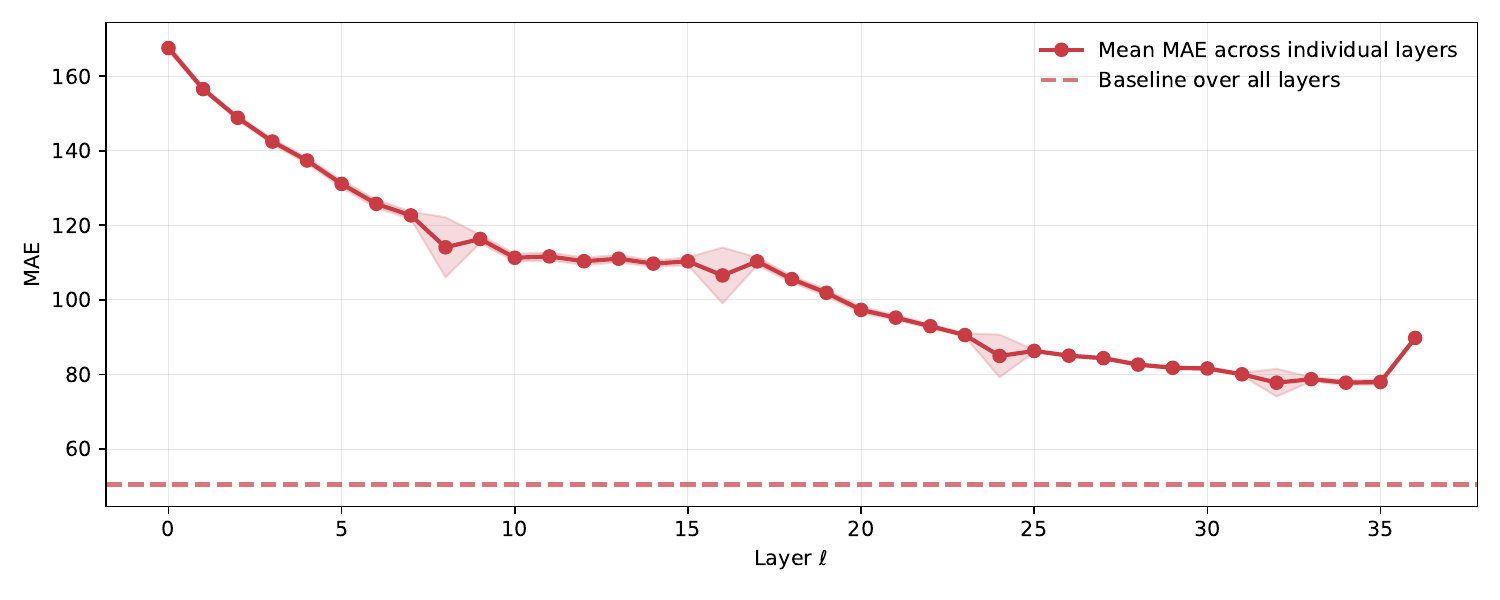}
\caption{Per-layer MAE of the Remaining Count Probe, in tokens (lower is better). Layer 0 is the embedding output and layer $L$ the final hidden state. The dashed line is the all-layers probe (Qwen3-4B / GSM8K).}
\label{fig:per-layer}
\end{figure}

We train a separate Remaining Count Probe on the state at each layer $\ell \in \{0, \ldots, L\}$ (Figure~\ref{fig:per-layer}). Error is high on early layers, falls through the middle of the network, and levels off in the upper third around $80$. The length information is therefore largely in place by the middle of the network: the bulk of the drop happens through the middle third, and accuracy does not improve further in the upper third. The probe at layer $0$, the embedding output, is far from the all-layers probe, so the signal is not a property of the current token in isolation. The all-layers probe used in the main text has lower MAE than every single-layer probe. \citet{shahout2024trail} report intermediate layers 8 to 15 as most informative; our sweep is on one model and a linear readout, so we read layer localization as dependent on the setup.

\section{Controls}
\label{app:controls}

\paragraph{Surface-feature ridge.} On Llama-3.1-8B / GSM8K, a ridge regression predicts $T$ from prompt surface features alone: prompt length in tokens, characters and words, and a TF-IDF bag of words over a $5{,}000$-term vocabulary. It is fit on the train split, with the regularization strength chosen by MAE on a held-out slice of it, and scored on naturally terminated eval examples, the protocol of the Completion Length Probe. The full surface model reaches $50.5$ against the probe's $42.3$ and the constant-median baseline's $58.7$, an improvement of $8.2$ tokens over the baseline against the probe's $16.4$. Length-only and TF-IDF-only variants reach $52.4$ and $52.0$, so the two signals mostly overlap. A static prompt predictor cannot produce any of the per-position results, having no state to update.

\paragraph{Shuffled-label controls.} Following \citet{hewitt2019control}, two probe heads share the Remaining Count Probe's architecture, input and training budget but are trained on permuted targets: a global shuffle across the batch, and a shuffle within each completion that preserves its set of targets. Scored against the true targets in the same run, they reach $66.5$ and $67.6$, against $34.5$ for the probe trained in that run and $70.2$ for the constant-median baseline.\footnote{The global control is slightly below the baseline because shuffling within a batch leaks some per-completion mean signal.} The within-completion control reaches its theoretical optimum, predicting each completion's mean remaining count, so the probe's advantage is tracking by position and not knowledge of each completion's overall length.

\paragraph{Resampling spread.}
\label{app:sampling-floor}
Under the stochastic decoding used for all data (temperature $0.7$, top-$p$ $0.8$, top-$k$ $20$), $T$ is a random variable given the prompt while the prompt's last hidden state is fixed. We drew $10$ fresh completions per prompt on the full GSM8K test set ($1{,}319$ prompts, Llama-3.1-8B; $0.41\%$ of draws reached the cutoff, consistent with the eval exclusion rate). Scoring each draw against the median of the other nine, the setting of a predictor that must guess an independent draw, gives a mean per-prompt spread of $34.8$ tokens (median $28.4$); scoring against the median of all ten gives $30.0$. The true spread lies between the two, so the probe's prompt-end error of $42.29$ is $1.22$ to $1.41$ times it. Of the range between the constant-median baseline ($58.66$) and the leave-one-out spread, the probe captures $69\%$ and the surface ridge $34\%$. The spread belongs to this decoding policy (it is zero under greedy decoding), which is the policy that generated the data, and it applies to prompt-end prediction only.

\begin{figure}[h]
\centering
\includegraphics[width=0.55\linewidth,trim=0 0 0 35.1,clip]{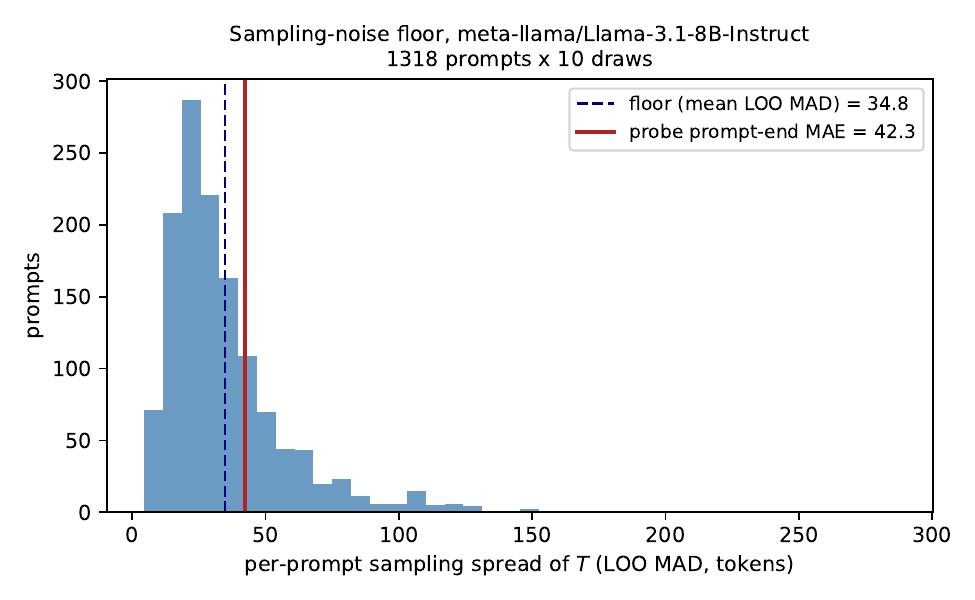}
\caption{Per-prompt spread of $T$ across $10$ draws (leave-one-out absolute deviation from the median, in tokens) on the GSM8K test set with Llama-3.1-8B. The probe's prompt-end MAE ($42.29$) is $1.22$ times the mean leave-one-out spread ($34.78$).}
\label{fig:sampling-floor}
\end{figure}

\paragraph{Scale.} On Qwen3-32B / GSM8K the full pipeline (extraction, probe training, evaluation) was run with one seed. The ratios to the constant-median baseline quoted in Section~\ref{sec:results} compare models from different families, so they show that the signal is present at 32B and are not a scaling trend.

\section{Transfer}
\label{app:transfer}
\label{app:cross-eval-floor}

Tables~\ref{tab:mistral-cross-abs} and~\ref{tab:olmo-cross-abs} give the Mistral-7B and Olmo-3-7B matrices, and Table~\ref{tab:cross-llama-evalfloor} divides each Llama entry by its eval-median floor. The closest call in the pooled count is Olmo OpenThoughts-1k to TriviaQA ($150.73$ against $151.00$); flipping it gives $28$ of $44$ and $p \approx 1.0 \times 10^{-6}$. Per-seed variation on every multi-seed Llama row is small (largest standard deviation $4.5$ tokens, GSM8K to Countdown; usually below $1$), so no cell of Table~\ref{tab:cross-llama-evalfloor} crosses $1$ under seed noise. The MMLU-Pro-trained row has one seed, and its borderline cell (to Countdown, $118.96$ against $117.82$) counts against the asymmetry.

\begin{table}[h!]
\centering
\small
\caption{Cross-dataset per-token MAE of the Remaining Count Probe on Mistral-7B, in tokens (lower is better), mean $\pm$ standard deviation over three seeds. The eval-median floor is defined as in Table~\ref{tab:cross-llama-abs}. Four training sets and five evaluation sets; MATH is omitted from this matrix and TriviaQA was not run.}
\label{tab:mistral-cross-abs}
\resizebox{\linewidth}{!}{%
\begin{tabular}{l rrrrr}
\toprule
Train & Count & Countdown & GSM8K & MMLU-Pro & OpenThoughts-1k \\
\midrule
\textit{eval-median floor} & 197.45 & 200.23 & 91.67 & 171.10 & 179.71 \\
\midrule
Count                  & 71.75 $\pm$ 0.18 & 188.31 $\pm$ 1.00 & 120.75 $\pm$ 0.05 & 229.31 $\pm$ 0.39 & 262.99 $\pm$ 0.35 \\
Countdown              & 257.66 $\pm$ 0.38 & 21.09 $\pm$ 0.09 & 124.40 $\pm$ 0.20 & 208.53 $\pm$ 0.22 & 226.64 $\pm$ 0.22 \\
\midrule
GSM8K                  & 247.25 $\pm$ 1.55 & 277.57 $\pm$ 1.48 & 60.93 $\pm$ 0.52 & 142.28 $\pm$ 1.21 & 170.10 $\pm$ 1.27 \\
OpenThoughts-1k        & 181.93 $\pm$ 0.89 & 219.30 $\pm$ 0.37 & 139.60 $\pm$ 4.57 & 128.60 $\pm$ 0.15 & 132.99 $\pm$ 0.35 \\
\bottomrule
\end{tabular}%
}
\end{table}

\begin{table}[h!]
\centering
\scriptsize
\caption{Cross-dataset per-token MAE of the Remaining Count Probe on Olmo-3-7B, in tokens (lower is better), mean $\pm$ standard deviation over three seeds. The eval-median floor is defined as in Table~\ref{tab:cross-llama-abs}.}
\label{tab:olmo-cross-abs}
\resizebox{\linewidth}{!}{%
\begin{tabular}{l rrrrrrr}
\toprule
Train & Count & Countdown & GSM8K & MATH & MMLU-Pro & OpenThoughts-1k & TriviaQA \\
\midrule
\textit{eval-median floor} & 116.54 & 117.84 & 124.32 & 178.83 & 184.82 & 219.01 & 151.00 \\
\midrule
Count                  & 28.97 $\pm$ 0.04 & 119.08 $\pm$ 0.17 & 169.72 $\pm$ 0.61 & 258.82 $\pm$ 0.77 & 271.78 $\pm$ 0.50 & 310.52 $\pm$ 0.63 & 185.82 $\pm$ 0.34 \\
Countdown              & 169.18 $\pm$ 0.19 & 4.57 $\pm$ 0.15 & 185.80 $\pm$ 0.19 & 284.90 $\pm$ 0.18 & 292.47 $\pm$ 0.18 & 339.74 $\pm$ 0.19 & 191.86 $\pm$ 0.18 \\
\midrule
GSM8K                  & 103.51 $\pm$ 0.65 & 115.47 $\pm$ 1.01 & 80.95 $\pm$ 0.13 & 147.89 $\pm$ 0.32 & 174.40 $\pm$ 0.35 & 203.76 $\pm$ 0.23 & 135.60 $\pm$ 0.07 \\
MMLU-Pro               & 125.74 $\pm$ 0.38 & 108.26 $\pm$ 0.11 & 130.51 $\pm$ 2.58 & 133.27 $\pm$ 0.22 & 134.73 $\pm$ 0.48 & 174.52 $\pm$ 0.30 & 158.45 $\pm$ 2.37 \\
OpenThoughts-1k        & 106.65 $\pm$ 0.29 & 104.97 $\pm$ 0.30 & 129.57 $\pm$ 3.11 & 130.81 $\pm$ 2.21 & 150.49 $\pm$ 0.75 & 178.03 $\pm$ 0.79 & 150.73 $\pm$ 0.70 \\
TriviaQA               & 110.91 $\pm$ 2.35 & 124.37 $\pm$ 2.42 & 102.90 $\pm$ 1.17 & 174.14 $\pm$ 4.61 & 164.93 $\pm$ 3.32 & 220.76 $\pm$ 4.18 & 117.25 $\pm$ 0.36 \\
\bottomrule
\end{tabular}%
}
\end{table}

\begin{table}[h]
\centering
\small
\caption{Table~\ref{tab:cross-llama-abs} (Llama-3.1-8B) with each entry divided by the eval-median floor of its column. Below $1$ beats the evaluation dataset's own constant-median baseline; bold marks off-diagonal cells below $1$.}
\label{tab:cross-llama-evalfloor}
\begin{tabular}{l rr | rrr}
\toprule
Train & Count & Countdown & GSM8K & MMLU-Pro & OpenThoughts-1k \\
\midrule
Count           & 0.30 & \textbf{0.84} & 1.44 & 1.33 & 1.49 \\
Countdown       & 1.56 & 0.04 & 1.25 & 1.37 & 1.48 \\
\midrule
GSM8K           & 1.11 & 1.27 & 0.55 & \textbf{0.95} & 1.06 \\
MMLU-Pro        & \textbf{0.89} & 1.01 & \textbf{0.91} & 0.70 & \textbf{0.78} \\
OpenThoughts-1k & \textbf{0.85} & \textbf{0.86} & 1.54 & \textbf{0.77} & 0.73 \\
\bottomrule
\end{tabular}
\end{table}

\paragraph{A regularization reading.}
\label{app:asymmetry-discussion}
The synthetic sets tie $T$ to a single deterministic feature of the prompt, so a linear probe fit on them can use any direction in $h_t$ correlated with that feature, and that direction need not carry length information on other data. The natural-language sets require a direction that explains length across many prompt structures, and that direction also covers the synthetic cases. The diagonal is the best entry in every row, but the off-diagonal pattern is inconsistent with a probe that has only memorized its training set's distribution of $T$. Tokenization is an unlikely primary explanation: a difference in tokens per character predicts failures in both directions, while the matrix shows transfer in one direction only. A replication with a shared tokenizer is left to future work.

\section{Retraction analysis}
\label{app:retraction}
\label{app:retraction-aggregate}

\paragraph{Curated example.} Table~\ref{tab:retraction-tokens} gives the tokens and predictions around the retraction in Figure~\ref{fig:retraction}a. The example was found by sorting eval completions by per-completion MAE and is an illustration only.

\begin{table}[h]
\centering
\footnotesize
\setlength{\tabcolsep}{4pt}
\caption{Tokens, true remaining count $r_t$ and Remaining Count Probe prediction $\hat{r}_t$ around the retraction of Figure~\ref{fig:retraction}a. The retraction tokens are positions 184 and 185.}
\label{tab:retraction-tokens}
\begin{tabular}{r l r r}
\toprule
$t$ & Token & $r_t$ & $\hat{r}_t$ \\
\midrule
179 & \texttt{"2"}     & 808 &   55.130 \\
180 & \texttt{"5"}     & 807 &    5.830 \\
181 & \texttt{" years"}& 806 &   19.612 \\
182 & \texttt{" old"}  & 805 &    7.057 \\
183 & \texttt{"**"}    & 804 &   39.003 \\
184 & \texttt{".\textbackslash n\textbackslash n"} & 803 &   71.384 \\
185 & \texttt{"Wait"}  & 802 &  277.406 \\
186 & \texttt{" \textemdash"}   & 801 &  255.722 \\
187 & \texttt{" that"} & 800 &  219.274 \\
188 & \texttt{" can"}  & 799 &  194.467 \\
189 & \texttt{"'t"}    & 798 &  217.419 \\
190 & \texttt{" be"}   & 797 &  208.706 \\
191 & \texttt{" right"}& 796 &  270.573 \\
\bottomrule
\end{tabular}
\end{table}

\paragraph{Token scan.} For every decoded token type in the Qwen3-4B / GSM8K eval completions, we computed the mean one-step change $\Delta\hat{r}_t$ at that token. ``Wait'' ranks first of $7{,}938$ token types (mean $+72.8$, $n = 55$, $95\%$ upward), against a mean change of $-0.93$ per token over all positions (the ground truth decrements by $1$). The scan ran on the same completions as the test below, so it does not provide an independent selection of the trigger.

\paragraph{Matched controls.} Each of the $55$ events is compared with control tokens from other completions in the same stratum: position within the completion in 100-token buckets, and per-completion mean-MAE quartile. Each event receives $100$ controls, $5{,}500$ in all. Positions within 10 tokens of any trigger in their own completion are excluded from the control pool, because the jump persists for several tokens. No stratum needed fallback sampling.

\paragraph{Statistics.} Retraction deltas have mean $+72.8$, median $+73.0$ and $95\%$ upward; controls have mean $-0.7$, median $-1.1$ and $48\%$ upward; Cohen's $d = 3.10$. The primary test is a sign-flip permutation over per-completion paired differences ($p < 10^{-4}$), since deltas within a completion are not independent. A stratified label shuffle over pooled events gives $p \approx 5 \times 10^{-4}$. By MAE quartile, $2$ events fall in the third and $53$ in the fourth; within the fourth, against controls from it, $d = 3.12$.

\section{Steering sweeps}
\label{app:steering-tables}

Table~\ref{tab:steering-all} gives the full sweep at each layer (Llama-3.1-8B / GSM8K, $n = 1{,}319$ paired prompts per arm). Every arm differs from $\alpha = 0$ at paired sign-flip $p < 10^{-4}$, including the degenerate arms, where generations run to the cutoff or the NLL rises. At layer 8, $\alpha = +4$ shortens completions and $\alpha = -12$ gives longer ones than $\alpha = -8$; at layer 16, $\alpha = -12$ gives longer completions than $\alpha = -8$, and $\alpha = +4$ already sends $67.1\%$ of generations to the cutoff.

\begin{table}[h]
\centering
\scriptsize
\caption{Full steering sweeps (Llama-3.1-8B / GSM8K, $n = 1{,}319$). Length ratio is the geometric mean of per-prompt length ratios against $\alpha = 0$. NLL is the mean per-token negative log-likelihood of the steered completion under the unsteered model (lower is better). At cutoff is the fraction of generations reaching $1{,}024$ tokens; the $0.5\%$ at $\alpha = 0$ is the natural rate.}
\label{tab:steering-all}
\begin{tabular}{ll rrrrrrr}
\toprule
Layer & & $\alpha = -12$ & $-8$ & $-4$ & $0$ & $+4$ & $+8$ & $+12$ \\
\midrule
\multirow{4}{*}{8}
 & mean length      & 44.2 & 21.7 & 144.9 & 231.9 & 201.9 & 861.8 & 1024.0 \\
 & length ratio     & 0.17 & 0.09 & 0.60 & 1.00 & 0.83 & 2.98 & 4.71 \\
 & NLL (unsteered)  & 5.13 & 3.36 & 0.58 & 0.18 & 0.32 & 0.52 & 0.24 \\
 & at cutoff (\%)   & 0.0 & 0.0 & 0.6 & 0.5 & 0.8 & 82.5 & 100.0 \\
\midrule
\multirow{4}{*}{16}
 & mean length      & 318.4 & 58.4 & 165.6 & 229.5 & 798.4 & 1023.4 & 1024.0 \\
 & length ratio     & 0.41 & 0.23 & 0.71 & 1.00 & 3.25 & 4.77 & 4.78 \\
 & NLL (unsteered)  & 3.38 & 1.74 & 0.30 & 0.18 & 0.29 & 0.27 & 0.27 \\
 & at cutoff (\%)   & 25.3 & 0.2 & 0.0 & 0.5 & 67.1 & 99.9 & 100.0 \\
\midrule
\multirow{4}{*}{24}
 & mean length      & 110.6 & 155.3 & 212.7 & 227.1 & 240.0 & 282.9 & 741.2 \\
 & length ratio     & 0.39 & 0.65 & 0.94 & 1.00 & 1.05 & 1.21 & 2.99 \\
 & NLL (unsteered)  & 0.57 & 0.31 & 0.20 & 0.18 & 0.19 & 0.26 & 0.33 \\
 & at cutoff (\%)   & 0.0 & 0.0 & 0.2 & 0.5 & 0.7 & 2.1 & 58.0 \\
\bottomrule
\end{tabular}
\end{table}

\section{Extended limitations}
\label{app:limitations-full}

\paragraph{Usability is not use.} Probes show what is available in the residual stream, and steering shows that an additive intervention along the decoded direction controls length. Neither shows that the model's unperturbed computation consults the direction; that needs activation patching or ablation of natural activity during ordinary generation. Our claims are about representation and about the direction's effect under intervention, not about the mechanism of normal decoding. Without a random-direction control, we have not shown that the length effect is specific to the probe direction, though the sign symmetry at layer 24 (negative $\alpha$ shortens, positive lengthens) is consistent with it.

\paragraph{Scope of the retraction result.} The aggregate test is on one cell, Qwen3-4B / GSM8K, because the stronger models rarely retract on this dataset; harder datasets would allow the same test on the primary models. Only ``Wait'' was tested. The matching conditions on position and MAE quartile but not on token identity, so the test does not separate an estimate that rises because more work follows from a probe weight on the ``Wait'' token, which in training precedes long remainders. The curated example comes from the worst-MAE region, where absolute predictions are far from the truth. Recovering the jump on completions where absolute predictions also track $r_t$ remains open.

\paragraph{Model scale and training regime.} The primary grid is 7--8B instruction-tuned models, with one-seed checks at 4B and 32B. Base models, whose length distributions differ, and frontier-scale models are untested. The per-layer sweep covers one model (Qwen3-4B) and does not says which layers matter in larger models.

\paragraph{Grid coverage.} Mistral-7B / TriviaQA was not run, and the Llama and Mistral transfer matrices cover five datasets (the Olmo matrix covers seven).

\paragraph{Natural termination.} On a generation that reaches the cutoff, $T$ is unobserved and $r_t$ is undefined, so any finite budget censors some sequences. On the long-form sets the filter is not negligible (Table~\ref{tab:exclusion}), and results there are conditional on termination. They do not address whether the model can tell that its generation will overrun.

\paragraph{A central tendency.} The Completion Length Probe estimates a central tendency of the length distribution, and much of its error is sampling noise under the decoding policy that generated the data (Appendix~\ref{app:sampling-floor}).

\end{document}